\documentclass[journal,transmag]{IEEEtran}
\usepackage{hyperref,color,breqn,multirow}
\usepackage[top=0.7in, left=0.65in]{geometry}
\usepackage{color}
\usepackage{graphicx}
\usepackage[T1]{fontenc}
\usepackage{array}
\usepackage{float}
\usepackage{multirow}
\usepackage{amssymb}
\usepackage{graphicx}
\usepackage{babel}

\makeatletter

\providecommand{\tabularnewline}{\\}
\floatstyle{ruled}
\newfloat{algorithm}{tbp}{loa}
\providecommand{\algorithmname}{Algorithm}
\floatname{algorithm}{\protect\algorithmname}

\makeatother

\begin{document}
\title{\fontsize{17pt}{17pt}\selectfont Boundary-Decoder network for inverse prediction of capacitor electrostatic
analysis}

\author{\IEEEauthorblockN{Kart-Leong Lim,
Rahul Dutta, and
Mihai Rotaru}
\vspace{5pt}

\IEEEauthorblockA{Institute of Microelectronics (A*STAR), Singapore \{limkl, duttar, mihai_dragos_rotaru\}@ime.a-star.edu.sg }}

\IEEEtitleabstractindextext{%
\begin{abstract}
Traditional electrostatic simulation are meshed-based methods which
convert partial differential equations into an algebraic system of
equations and their solutions are approximated through numerical methods.
These methods are time consuming and any changes in their initial
or boundary conditions will require solving the numerical problem
again. Newer computational methods such as the physics informed neural
net (PINN) similarly require re-training when boundary conditions
changes. In this work, we propose an end-to-end deep learning approach
to model parameter changes to the boundary conditions. The proposed
method is demonstrated on the test problem of a long air-filled capacitor
structure. The proposed approach is compared to plain vanilla deep
learning (NN) and PINN. It is shown that our method can significantly
outperform both NN and PINN under dynamic boundary condition as well
as retaining its full capability as a forward model.
\end{abstract}

}

\maketitle
\thispagestyle{empty}
\pagestyle{empty}

\section{Introduction}

The test case chosen to demonstrate the proposed approach is a simple
electrostatic problem of a rectangular coaxial capacitor (Fig 1) [1], [3].
The electrostatic problem can be formulated as a Laplace equation
which can be then solved applying FDM with a successive over-relaxation
(SOR) algorithm [2], [3]. Although SOR is
accurate and easy to implement, it is slow to compute as it has to
comb through each node for the entire 2D grid to perform iterative
update. Also, when the boundary conditions changes the computation
has to be repeated. In this work, we target how to allow parameter
changes on a single boundary condition without incurring new computational
cost.

Considering the problem sketched in Fig 1, this structure has a symmetry
that can be exploited when the origin is placed in the middle of the
inner plate. With this arrangement only a quadrant of the problem
need to be considered. For different lengths of the middle plate the
parameter $d$ changes which affects the boundary conditions, which
in turns affects the solution of this problem. The effects of the
electric potential distribution $V(x,y)$ with changes in $d$ are
shown graphically in Fig 2 (Top). The boundary conditions definition
as used in the FDM are given in eqn (1). In Fig 1 $a,b$ are assumed
fixed. The effect of increasing $d$ on $V$ is such that as $d\in[0,1]$
increases, $V$ increases in volume as shown in Fig 2 (top row).
\begin{equation}
\begin{array}{c}
V=\frac{\delta V}{\delta y},\;\frac{d}{2}\leq x\leq\frac{a}{2},\;y=0\\
V=V_{0},\;1\leq x\leq\frac{d}{2},\;y=1
\end{array}
\end{equation}

\begin{figure}
\begin{centering}
\includegraphics[scale=0.7]{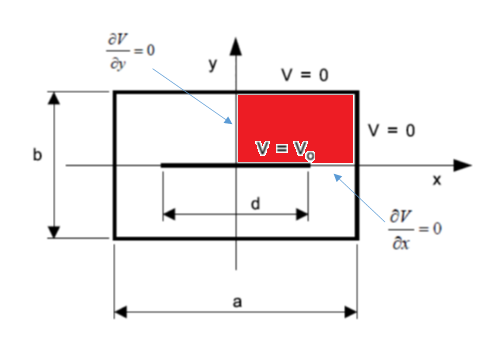}
\par\end{centering}
\caption{Boundary conditions of an air-filled capacitor. Parameter 'd' is allowed
to vary. Only the first quadrant (shaded red) of the distribution
$V(x,y)$ needs to be solved due to symmetry.}
\end{figure}

\begin{figure}
\begin{raggedright}
\includegraphics[scale=0.44]{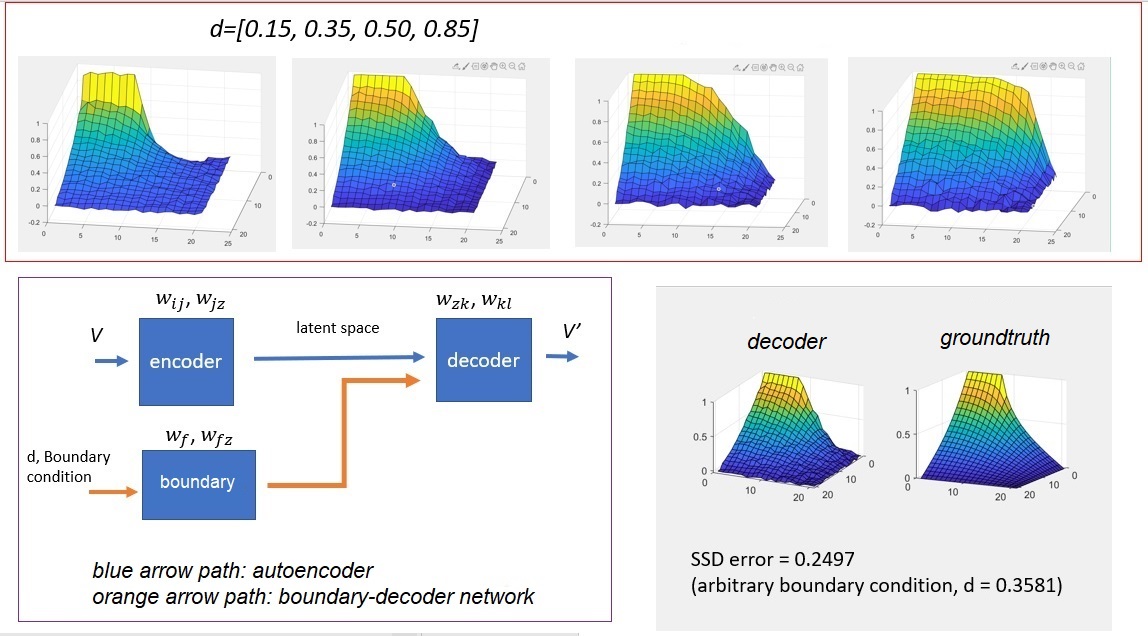}
\par\end{raggedright}
\caption{(Top) $V'$ plots at decoder corresponding to varying $d$. (Lower
left) Inverse prediction using proposed method. (Lower right) Reconstructed
$V'$ vs groundtruth (SOR) and the sum of squared error.}
\end{figure}

\subsection{Problem Statement}

Given that a regression model can model $1...m$ samples of electrostatic
field as $V_{1...m}$ and corresponding boundary condition parameter
as output $BC\left(V_{1...m}\right)$. Inverse prediction then tries
to recover input $\hat{V}$ when presented with an output $BC\left(\cdot\right)=d$.

Inverse prediction is an ill-posed problem due to: (i) The dimension
of input is much larger than the output i.e. $\hat{V}\in\mathbb{R}^{1\times N}$
and $BC(\hat{V})\in\mathbb{R}^{1}$ respectively. (ii) The regression
model $\phi$ in Algorithm 1 is sparsely distributed with a large
number of zero coefficients. (iii) An initial estimate is necessary.
Due to (i), (ii) and (iii), the optimization of $\hat{V}$ involves
root solving a large number of coefficients e.g. $N=16081$. Even
a good initial value does not always guarantee a good inverse prediction.

To reduce the problems of (i), (ii) and (iii), we can use an autoencoder
to model an infinite number of electrostatic field simulations with
different parameter changes in the latent space. However, to compute
inverse prediction in the latent space, the computation becomes a
long-winded problem in \textbf{Algorithm 1}: offline training autoencoder
via encoder-decoder loss\footnote{$T$ and $\hat{\gamma}$ are the target and network output respectively.}
using (1a) followed by offline training regression model using (1b)
in the latent space. Then, online inverse prediction using (1c) in
the latent space followed by lastly online $V(x,y)$ decoder reconstruction
using (1d). 

\section{Proposed Method}

\begin{algorithm}
\caption{Baseline inverse prediction.}
\begin{tabular}{cc}
1a) enc-dec & $\min_{w}\left\Vert T-\hat{\gamma}\right\Vert ^{2}$\tabularnewline
1b) regress model & $\left[z_{1}\cdots z_{m}\right]\cdot\phi=\left[d_{1}\cdots d_{m}\right]$\tabularnewline
1c) inverse pred. & $\begin{array}{c}
\arg\min_{\hat{z}}\;\left(\hat{z}\cdot\phi-d\right)\end{array}$\tabularnewline
1d) reconst. & $\begin{array}{c}
\hat{V}=\tanh\left(\sum_{k}w_{kl}\cdot\tanh\left(\sum_{z}w_{zk}\cdot\hat{z}\right)\right)\end{array}$\tabularnewline
\end{tabular}
\end{algorithm}

\begin{algorithm}
\caption{Proposed inverse prediction}
\begin{tabular}{ccc}
2a) enc-dec+ & \multirow{2}{*}{$\min_{w}\left\Vert T-\hat{\gamma}\right\Vert ^{2}-\lambda\left\Vert \gamma-\hat{\gamma}\right\Vert ^{2}$} & \tabularnewline
bou-dec &  & \tabularnewline
\multirow{1}{*}{2b) reconst.} & $\begin{array}{c}
\hat{V}=\tanh\left(\sum_{k}w_{kl}\cdot\tanh\left(\sum_{z}w_{zk}\cdots\right.\right.\\
\left.\cdots\times\sum_{f}w_{fz}\cdot\sum w_{f}\cdot d\right)
\end{array}$ & \tabularnewline
\end{tabular}
\end{algorithm}

We simplify inverse prediction by proposing an end-to-end approach
known as the boundary-decoder network in \textbf{Algorithm 2}. Both
regression and inverse prediction are replaced by the boundary network.
Instead, we directly feed an arbitrary boundary parameter to the network,
and it will online reconstruct a corresponding $V$ at the decoder
using (2b). The offline training only involves a boundary-decoder
loss\footnote{$\gamma$ and $\hat{\gamma}$ are the groundtruth and network output
respectively.} in (2a) using the architecture in Fig 2 (bottom left). The advantages
are: (i) \textbf{Efficient representation}: regression modeling, inverse
prediction and latent representation are all handled by a boundary
network. (ii) \textbf{Semi-supervised model}: the latent space is
jointly trained by both encoder-decoder (unsupervised) and boundary-decoder
network (supervised).

\section{Experiments}

In Table I, we compare the sum of squared error of reconstructed $V$
vs groundtruth (SOR) using baseline and proposed method. Without any
latent space, the overall inverse prediction is the poorest. Also
for inverse prediction, we use a less accurate initial estimate to
be $\pm0.2$ from the actual $d$. For baseline we only train the
encoder-decoder network (unsupervised). For proposed method, we separately
train the boundary-decoder network (supervised) and the encoder-decoder+boundary-decoder
network (semi-supervised). From the result, we observed that inverse
prediction using the unsupervised method give the worst result. The
proposed method does not require any initial estimate. Although the
size of supervised sample is smaller than unsupervised, the proposed
boundary-decoder network alone already outperforms the baseline. When
we allow the latent space to be updated by both encoder-decoder and
boundary-decoder network, the proposed semi-supervised method achieve
the lowest error. ``enc-dec'' refers to encoder-decoder network
and ``bou-dec'' refers to boundary-decoder network.

\begin{table}
\caption{Sum of squared error comparison}
\begin{tabular}{|c|c|c|c|c|c|c|}
\hline 
loss func. & d=0.3 & d=0.4 & d=0.5 & d=0.6 & d=0.7 & d=0.8\tabularnewline
\hline 
\hline 
raw space & 6.492 & 4.718 & 11.34 & \textbf{0.196} & 3.688 & 2.733\tabularnewline
\hline 
enc-dec & 5.557 & 3.291 & 1.868 & 3.962 & 1.606 & 3.789\tabularnewline
\hline 
bou-dec & 0.491 & 0.741 & 1.518 & 1.201 & 0.471 & 0.086\tabularnewline
\hline 
enc-dec+bou-dec & \textbf{0.335} & \textbf{0.215} & \textbf{0.661} & 0.642 & \textbf{0.277} & \textbf{0.063}\tabularnewline
\hline 
\end{tabular}
\end{table}

In Table II, we compare our enc-dec+bou-dec approach with other deep
learning methods. Both plain vanilla neural network (NN) and physics
informed neural net (PINN) [4] can only train
on fixed boundary condition. We train both methods on $d=0.55$ but
we compare how well they perform when $d$ deviates from training.
We observed that PINN performs better than NN since it can benefit
from PDE loss. But both methods are significantly poorer than the
proposed method.

\begin{table}
\caption{Comparison with state-of-the-art}

\centering{}%
\begin{tabular}{|c|c|c|c|}
\hline 
 & d=0.6 & d=0.65 & d=0.7\tabularnewline
\hline 
NN & 3.307 & 4.874 & 8.293\tabularnewline
\hline 
PINN & 3.0 & 4.541 & 6.835\tabularnewline
\hline 
enc-dec+bou-dec & \textbf{0.642} & \textbf{0.8127} & \textbf{0.277}\tabularnewline
\hline 
\end{tabular}
\end{table}

\section{Conclusion}
 We propose the boundary-decoder network which is an end-to-end approach that allows a user to directly feed boundary information into the latent space such that the decoder can reconstruct an electrostatic distribution that accurately corresponds to the boundary. We demonstrated our proposed method on a long air-filled capacitor dataset and compared that to using other computational method such as the PINN and the NN found in electrostatic analysis. We showed that our method can easily outperforms both NN and PINN under dynamic boundary condition of the underlying governing Laplace equation, as well as regression model both in the original space and latent space.

\end{document}